\documentclass[dvipsnames]{article}
\usepackage{custom}

\usepackage{lineno}

\usepackage{graphicx}
\usepackage{amsmath}
\usepackage{amssymb}
\usepackage{booktabs}
\usepackage{multirow}
\usepackage{comment}
\usepackage[utf8]{inputenc}                                 %
\usepackage[australian]{babel}                              %
\usepackage[T1]{fontenc}                                    %

\usepackage[activate={true,nocompatibility},final,tracking=true,kerning=true,spacing=true,factor=1100,stretch=10,shrink=10]{microtype}
\microtypecontext{spacing=nonfrench}

\usepackage[inline]{enumitem}

\usepackage[autostyle=try,                      %
  style=australian]{csquotes}
  \MakeOuterQuote{"}                    %

\usepackage{amssymb}                                        %
\usepackage{amsmath}
\usepackage{braket}                                         %

\usepackage{graphicx}                                       %

\usepackage{tikz}

\usepackage{subcaption} %

\usepackage{booktabs}
\usepackage{placeins}

\usepackage{xcolor}

\usepackage[boxed,ruled,vlined,linesnumbered]{algorithm2e}
\DontPrintSemicolon
\SetKwProg{Fn}{function}{}{}
\SetKwFunction{FnFeatureExtraction}{FeatureExtraction}
\SetKwFunction{FnPoiNet}{PoiNet}
\SetKwFunction{FnBoundingBoxGenerator}{BoundingBoxGenerator}
\SetKwFunction{FnActivityNet}{ActivityNet}
\SetKwFunction{FnSampleFree}{SampleFree}
\SetKwFunction{FnRestartArm}{RestartArm}
\SetKwFunction{FnPickArm}{PickArm}
\SetKwFunction{FnRewire}{Rewire}
\SetKwFunction{FnNearest}{Nearest}
\SetKwFunction{FnRestartArm}{RestartArm}
\SetKwComment{Comment}{$\triangleright$\ }{}
\SetKwInput{KwInit}{Initialise}
\SetKwInOut{KwPara}{Parameter}

\SetCommentSty{mycommfont}

\usepackage{cleveref}                                        %
\crefname{assumption}{assumption}{assumptions}
\crefname{problem}{problem}{problems}
\crefname{algorithm}{Alg.}{Algs.}
\Crefname{algorithm}{Algorithm}{Algorithms}
\crefname{figure}{Fig.}{Figs.} %
\crefformat{equation}{(#2#1#3)}
\crefrangeformat{equation}{(#3#1#4) to~(#5#2#6)}
\crefmultiformat{equation}{(#2#1#3)}%
{ and~(#2#1#3)}{, (#2#1#3)}{ and~(#2#1#3)}

\usepackage[
    backend=biber
  ,minnames=3                       %
  ,date=year                %
  ,maxbibnames=99
  ,labeldate=year
  ,style=authoryear
  ]{biblatex}
\AtEveryBibitem{                      %
  \iffieldequalstr{eprinttype}{jstor}
  {\clearfield{eprint}}
  {}
}

\DeclareSourcemap{
  \maps{
    \map{
      \pertype{inproceedings}
      \step[fieldset=publisher, null]
      \step[fieldset=urldate, null]
      \step[fieldset=volume, null]
      \step[fieldset=pages, null]
    }
  }
}
\renewrobustcmd*{\bibinitdelim}{\,} %
\Crefname{figure}{Fig.}{Figs.}

\begin{document}

\title{Transfer Learning for Assessing Heavy Metal Pollution in Seaports Sediments}

\author{Tin Lai%
\thanks{
School of Computer Science, The University of Sydney
}
\,\thanks{
tin.lai@sydney.edu.au
}%
\And
Farnaz Farid%
\thanks{
School of Social Sciences, Western Sydney University
}%
\And
Yueyang Kuan%
\textsuperscript{\normalfont $\dagger$}
\And
Xintian Zhang%
\textsuperscript{\normalfont $\dagger$}
}

\maketitle

\begin{abstract}
Detecting heavy metal pollution in soils and seaports is vital for regional environmental monitoring. The Pollution Load Index (PLI), an international standard, is commonly used to assess heavy metal containment. However, the conventional PLI assessment involves laborious procedures and data analysis of sediment samples. To address this challenge, we propose a deep-learning-based model that simplifies the heavy metal assessment process. Our model tackles the issue of data scarcity in the water-sediment domain, which is traditionally plagued by challenges in data collection and varying standards across nations. By leveraging transfer learning, we develop an accurate quantitative assessment method for predicting PLI. Our approach allows the transfer of learned features across domains with different sets of features. We evaluate our model using data from six major ports in New South Wales, Australia: Port Yamba, Port Newcastle, Port Jackson, Port Botany, Port Kembla, and Port Eden. The results demonstrate significantly lower Mean Absolute Error (MAE) and Mean Absolute Percentage Error (MAPE) of approximately 0.5 and 0.03, respectively, compared to other models. Our model performance is up to 2 orders of magnitude than other baseline models. Our proposed model offers an innovative, accessible, and cost-effective approach to predicting water quality, benefiting marine life conservation, aquaculture, and industrial pollution monitoring.
\end{abstract}

\newcommand{\orcidauthorA}{0000-0000-0000-000X} %

\section{Introduction}

\renewcommand{\cite}{\autocite}

Shipping is the backbone of international trade and the global economy. About 80\% of the world's trade volume and 70\% of trade value is generated by sea \cite{sirimanne2019review}. Some island and coastal states rely greatly on ports for global trade.
According to statistics, Australia depends almost entirely on sea transport for trade, and 98\% of Australia's goods are shipped through ports to all parts of the world. However, the impact of such a large number of ships puts an enormous weight on the marine environment.
As for transportation using fossil fuels, vessels emit large amounts of harmful gases, oil spills, marine waste, metal pollution, and other pollutants.
Several sources of water pollution around ports exist, including oil and gasoline spills from ships, paint leaching, metal contamination, and transfer of harmful aquatic organisms \cite{trozzi2000environmental}. Among them, metal pollutants are considered the most severe pollution due to their substantial toxicity, slow degradation~\cite{marshall2010isolating}, vast sources and rapid diffusion~\cite{mitra2022impact}. Metal pollutants bind to suspended particles in the water under certain conditions in the harbour environment and then form deposits on the bottom. The metal concentration in this sediment could effectively reflect the pollution level of the port over a period of time and is more accurate than directly detecting the pollution level by the water body. Therefore, sediments are essential evidence of human influence on the ocean.
As a result, many researchers use sediment analysis to examine transport~\cite{gong2014review} and environmental conditions~\cite{casado2009multivariate} in ports and seas.
However, detecting the water quality of the port directly through sediment analysis is very costly.
Due to the need to sample sediments in the target study waters, each heavy metal's content is obtained through a series of physical and chemical methods~\cite{xu2022critical}.
This process would consume a lot of time and money. This cost scales dramatically to cover the increasing logistic volume of international trade and the need to impose ports and maritime regulation~\cite{saliba2022shipping}.

In recent years, many machine learning models have been applied to solve various environmental engineering problems, like water quality prediction. For example, machine learning modelling is a straightforward solution for trade or logistic predictions when sufficient data is given for model training.
Although machine learning could solve practical problems very well in some circumstances, it often requires a relatively large amount of sample data to cover and learn the distribution of the population data.
In particular, marine sediment information, metal content measurement, and their corresponding impact are often hard to obtain due to difficulties such as time constraints, maritime regulations, and the effort it takes to collect sediment samples in the middle of an ocean. Therefore, this research focuses on using a small amount of data to build a machine-learning model capable of predicting port water quality with high prediction accuracy.

In this paper, we propose a transfer learning-based deep learning modelling architecture that predicts the pollution index of water sediment using only a limited amount of data.
Transfer learning modelling---a novel methodology that previous water quality researchers had yet to attempt---is shown to be a promising approach even with the presence of data sparsity issues.
Transfer learning consists of two components---source domain and target domain. The dataset used for the source domain is usually much larger than the dataset used for the target domain. In addition, it is also required that the tasks of the two fields contain shared mutual information, such that features from the source domain is helpful for the target domain.
This paper presents the modelling architecture and foundations introducing transfer learning as a suitable approach for marine environmental water quality prediction.

Section II introduces the background of the research, including the basic ideas of constructing datasets and the basis of transfer learning. Section III describes data collection and preprocessing. Section IV describes the structure of the DNN-based transfer model in detail. Results and comparisons of the experiments are provided in Section V. Section VI is the summary and discussion of the study. Finally, section VII outlines the limitations of the research and future expectations.
Our contributions are as follows.
\begin{enumerate}
\item We first present the fundamentals of transfer learning and our Deep Neural Network (DNN)-based model architecture for constructing a model capable of learning from some source domain for prediction in some target domain with a limited amount of data.
We use the PLI of water ports in NSW, Australia, as our case study.
\item We present an empirical study comparing our DNN transfer learning approach against traditional machine learning for water quality prediction.
In particular, we focus on comparing the predictive accuracy when the target domain only contains a limited amount of data.
\item Finally, we performed a parametric study to analyse the sensitivity of our model and present the model with the optimal set of parameters on the water-sediment data.
\end{enumerate}
In particular, we propose a DNN-based transfer learning model, compare its performance with traditional machine learning methods, and investigate the feasibility of applying transfer learning to water quality prediction.

\section{Literature review}
\subsection{Factors Affecting the Environment}
Features related to economic activities, ecology and geography can detect ports' water and air quality. The features of economic activities consist of population density, employee income, number of businesses, and number of registered motor vehicles. This study's ecological features include temperature, total rainfall, wind speed, and humidity. The geographic features consist of latitude and longitude.

For water quality, much research has used population densities as the driving factors to predict and quantify the land-based waste entering the ocean. Although population density does not directly affect water pollution, human-related activities such as land use, infrastructure, and socio-economic factors would increase the use of plastic, which affects the ocean's water quality \cite{schuyler2021human}. Secondly, one study points out that income also has a small direct impact on water pollution. Still, increased income represents economic growth, which would increase pollutant emissions \cite{borhan2012green}. Thirdly, with the development of industry, industrial facilities, sewage treatment plants and mining operations discharge wastewater into the ocean through rivers and lakes, which results in a large part of the world's water pollution sources. Wastewater discharged from industries and cities and chemicals would be released along drains and eventually flow into the ocean. According to \cite{parris2011impact}, chemical fertilisers and pesticides used in agricultural production flow into the sea through rivers. Pollution in agricultural production involves surface water, groundwater and seawater, severely impacting the entire marine ecosystem.

Forestry also impacts water quality, but it is generally regional and less intensive compared to activities related to agriculture or urban \cite{fulton2002forestry}. Fourthly, motor Vehicle emissions could affect water quality. During the manufacture of motor vehicles, Cadmium can enter water bodies through rain and cause pollution \cite{talebzadeh2021cadmium}. For ecological features, pH, biochemical oxygen demand (BOD), temperature, Ultraviolet (UV) index, turbidity, and conductivity represent the physical and chemical changes in water quality, which could be used as attributes to evaluate water quality \cite{dunca2018water}. In addition, rainfall is also a factor affecting water pollution observed in \cite{sinha2017eutrophication}.
Studies have shown that rainfall will wash the nitrogen and phosphorus produced by human activities into the rivers and lakes and eventually flow into the ocean. As a result, the output of harmful nitrogen will increase, and the water will become eutrophicated~\cite{modabberi2020caspian}. Moreover, wind speed is a factor affecting the degree of water eutrophication as well. Studies indicate that low wind speed will increase the phosphorus (P) and decrease the nitrogen (N) in sediment \cite{deng2018climatically}. Sallam and Elsayed's study shows that air temperature and humidity have been measured concerning the water quality of the lake \cite{sallam2015estimating}.
Regarding the geographic features, one study suggested a relationship between temperature variability and latitude and longitude has more influence on temperature than longitude \cite{shao2012characteristics}. Geographic Features indirectly affect water quality through the temperature.

Particulate matter 2.5 (PM2.5) measures the air quality, which refers to tiny particles or droplets in the air that are two and one-half microns or less in width. Population density is negatively correlated with air quality in a study by Brock and Schrauth \cite{borck2021population}. Additionally, the study has shown a correlation between income and air pollution levels. Low-income communities tend to have higher average pollutant levels \cite{finkelstein2003relation}. Factories discharge chemicals and pollutants into the water and air, so a linear correlation exists between energy consumption and air pollutants in factories \cite{wu2021multimodal}. Gasoline combustion by motor vehicles causes air pollution since motor vehicles emit hazardous substances such as PM, nitric oxide, carbon monoxide, organic compounds, and lead \cite{kjellstrom2006air}.
Regarding the ecological features, air temperature indirectly affects the movement of air pollution since air temperature directly affects the movement of air. For example, even though the industrial emissions remained unchanged throughout the year, the carbon monoxide pollutant increased from wood burning during the cold temperature condition. On the other hand, ground-level ozone pollution is produced during hot temperature conditions. Moreover, heavy rainfall has a negative impact on air quality \cite{kwak2017identifying}. Wind speed is a related factor in the diffusion of pollutants in the lower atmosphere layer \cite{finzi1984stochastic}. Humidity is the gaseous of water in the air, and high humidity increases the amount of harmful or toxic chemicals in the air, which could be one of the causes of air pollution. Geographic features also contribute to air quality, like water quality, since Geographic features directly affect the temperature.

\subsection{Transfer learning}
Machine Learning had shown promising results in numerous traditional fields like in robotics~\cite{lai2020bayeLocal} and financial sectors~\cite{forexNonStationaryTimeSeries}.
Inspired by the human ability to transfer knowledge across domains or disciplines, transfer learning aims to improve performance or minimise the data needed in the target domain with knowledge from the source domain. However, transfer learning is unsuitable for all new tasks because knowledge transfer would only succeed if the target and task domain contained sufficient standard features. For example, learning to swim does not help people learn to play Go any faster. However, we can often project the abstract features that we have learned from a related domain into other similar tasks, thereby improving our model's predictive power.
Such an approach is akin and can often be beneficial in other fields where data are often limited, like in the medical domain, where collected data from rare cases are highly limited~\cite{sensorsMLforDiabetes}.

Transfer learning is usually divided into Homogeneous transfer learning and Heterogeneous transfer learning \cite{zhuang2020comprehensive}. Inhomogeneous transfer learning, the feature space between domains overlaps and the label space is the same. Still, the marginal or predicted distributions are not the same between domains. In Heterogeneous transfer learning, the feature space between domains does not overlap, or the label space is entirely different. The survey by Fuzhen et al. \cite{zhuang2020comprehensive} introduces and summarises more than 40 common transfer learning methods, most of which belong to Homogeneous transfer learning. The survey by Day and Khoshgoftaar \cite{day2017survey} also reviews Heterogeneous transfer learning and finds substantial effects on their predictive power.
Transfer learning transfers common knowledge rather than task-specific knowledge between tasks.
The datasets of the two domains jointly determine the specific transfer learning method.
Transfer learning can only be used when the two tasks have similarities; otherwise, it may cause negative effects.

\section{Dataset preparation}
\subsection{Data Collection}
The data used in the research are about six years of environmental and economic data in parts of coastal areas in Australia. The data are divided into original domain datasets \cite{parker2017datalab} and target domain datasets \cite{jahan2018comparison}. The dataset of the two domains includes population density, total employee income, number of businesses, number of registered motor vehicles, water temperature, longitude, latitude, total rainfall, wind speed, and humidity. However, the difference between these two datasets is that the responses are different, the label of the former is PM2.5, and the latter is labelled Pollution Load Index (PLI).

\subsection{Data Analysis}
Assessing the risk of heavy metal can help to prevent catastrophic contamination of lakes or even reservoirs~\cite {aradpour2021alarming}. The source domain dataset used in this study is collected from the Australian Bureau of Statistics\cite{parker2017datalab}. The dataset contains environmental, economic and air quality data of some Australian cities (including Sydney, Wollongong, Newcastle, Canberra, Melbourne, Geelong, Adelaide, Darwin, Tasmania, Geraldton, Perth, and Mandurah) for the past six years. The target domain dataset collects 2017 environmental, economic and PLI data for the waters of six ports (Port Jackson, Port Botany, Port Kembla, Port Newcastle, Port Yamba, Port Eden) in New South Wales, Australia \cite{jahan2018comparison}. Moreover, PLI is calculated from the port sediments' metal content to assess the pollution degree.

The sampling interval of sample data in the same region is monthly. The reason is that considering environmental variables, such as temperature, humidity, wind speed and air quality, there is usually a balanced change from month to month compared to a sampling interval of days or quarters. Additionally, the sampling interval of quarter-month is more suitable for economic indicators because the statistical cycle of economic data is usually quarterly or yearly. After careful consideration, an area's data sampling interval is monthly, equivalent to thirty days.

\subsection{Data Pre-processing}
The original domain and target datasets have 1526 and 6 records, each containing 16 features and a label. Since not part of the features is counted quarterly or annually, a linear regression is performed on this part of the feature data within the year, and each month's value of these features has been estimated. In addition, due to the model does not involve timing problems, some features, including Date, Country, City, and Port, are deleted in the actual modelling, and Longitude, Latitude are used instead of Country, City, and Port.

\section{Methodology}

Our transfer learning-based DNN model provides a flexible learning procedure suitable for several seaport datasets.
Using a DNN-based model provides us with a flexible way to extract information from the seaport sediment data.

\subsection{Model Architecture}
\begin{figure}[tb]
    \centering
    \includegraphics[width=0.55\linewidth]{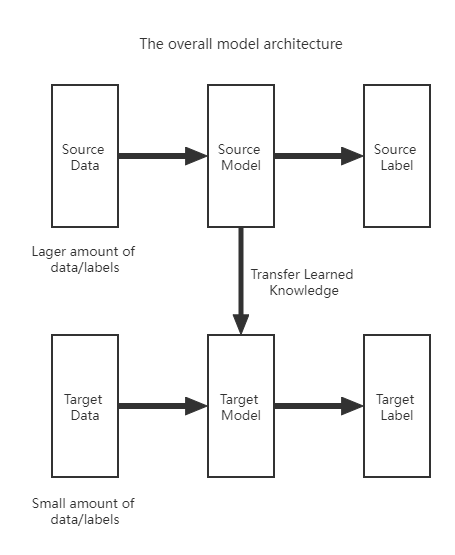}
    \caption{The overall model architecture of source and target domain
    \label{fig:overall}}
\end{figure}

Figure \ref{fig:overall} shows the overall model architecture. The model is divided into two parts, source domain and target domain. Firstly, use source data to train a source model, which is also called a pre-training model. For example, we use a pre-training model to predict the PM2.5 of major cities in Australia. Previously-trained knowledge is then parameterised and transferred to the target domain model. Afterwards, the target dataset was used to train the target domain model, which could be used to predict the PLI of the waters of various ports in NSW and Australian ports. Our model architecture provides a flexible approach to adapting our source dataset into different target domains. In the following section, we will elaborate on the source model, target model and transfer process in detail.

\subsubsection{Source Model}
\begin{figure}[tb]
    \centering
    \includegraphics[width=0.65\linewidth]{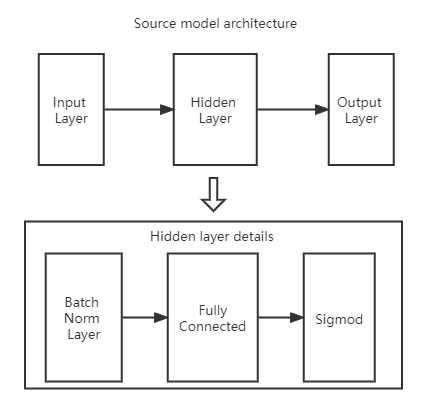}
    \caption{Source model architecture
    \label{fig:source-model}}
\end{figure}

We produce a source model by pre-training this model for predicting PM2.5 in regional Australia. Figure \ref{fig:source-model} illustrates the overall structure of the source model. The source model includes an input, hidden, and output layer. The hidden layer comprises batch norm, fully-connected, and sigmoid activity layers. The input layer has 12 nodes, the output layer has one node, and the size and number of hidden layers are adjusted according to the model's specific performance.

\subsubsection{Transfer Method}
Conservative training is used as a transfer learning method in this paper. We begin by freezing the pre-trained model's learned weight in the source domain.
The frozen parameters of the model would not be updated during our training procedure.
Then, we begin another training procedure to update the parameter values of the last fully connected layer.
We first randomise the weights of the last layer, followed by another model training procedure on the target domain. The dataset of the source and target domain determines the choice of this transfer learning method. Unfortunately, the amount of data in the source domain needs to be larger, and the data in the target domain needs to be bigger, resulting in many other transfer learning methods that could be more suitable for this research.

\subsubsection{Target Model}
\begin{figure}[tb]
    \centering
    \includegraphics[width=0.65\linewidth]{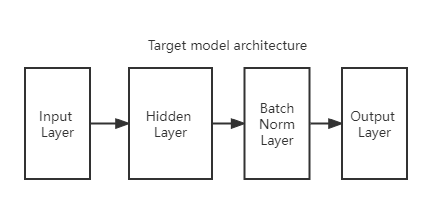}
    \caption{Target model architecture
    \label{fig:target-model}}
\end{figure}

The target model was used to predict PLI in port waters in NSW, Australia. As shown in Figure \ref{fig:target-model}, the difference from the pre-trained model is that the target model has an additional batch norm layer before the output layer.

\subsection{Evaluation Methods}
MAE and MAPE are used as the evaluation method of the regression task model in this research. The value of MAE ranges from 0 to infinity. MAE is equal to 0 when the predicted value is exactly the same as the true value, which means the model is perfect. The value of MAPE ranges from 0 to infinity, and MAPE is very similar to MAE. The smaller the value, the better the model. The model is classified as inferior when MAPE is greater than 100\%. The following two reasons were considered for using MAE and MAPE: (i) MAE and MAPE are better for the experiment that is regression task, and (ii) MAE and MAPE use the median calculation method for outliers, and the median is more stable than the mean for outliers \cite{chai2014root}.

\section{Result}
The experiments are divided into two parts, conducted in the source domain and task, respectively. The purpose of experiments in the source domain is to build the best pre-trained model for transfer learning, while experiments in the target domain focus on improving the final predictive capabilities.

There are several comparative experiments in the source domain, including the comparison of (i) the optimal method and the parameter set, (ii) the number of hidden layers and the size,  (iii) the weight initialisation method, (iv) the earning rate and (v) various model methods. Through these controlled trials, the model approach with the best performance on the source domain task and its configuration can be identified. However, the comparison model approach is the only experiment on the target domain.

\subsection{In Source Domain}
Use DNN as a pre-trained model for training, given the high predictive accuracy and transferability of DNN. The source domain dataset is used to train the model. The dataset contains 10 features, which the source domain model uses to predict PM2.5. This subsection describes the training process and parameter tuning of the pre-trained model.

\subsubsection{Base Model Sets}
The base model was established as a control group, and the parameter tuning of all subsequent experiments was performed on the base model. As shown in Table \ref{table:base-model}, the basic model is the DNN model that includes two hidden layers, and each hidden layer has 24 nodes. The base model adopts the optimisation method of Momentum SGD, and the learning rate is 0.01. The initial weight method of the model uses Xavier initialisation. Evaluation methods are MAE and MAPE.

\subsubsection{Optimizer and Parameters}
The optimiser is used to find the smallest loss after the loss function is determined. Therefore, the model could learn the correct knowledge in the iterative process only if the optimiser is used. Table \ref{table:compare-optimiser} presents a performance comparison of the model using five different optimisers. The result illustrates that models with Momentum Stochastic Gradient Descent (SGD) and ADAM have approximate MAE, but the former has a lower MAE, 0.1828.

Table \ref{table:compare-optimiser} has shown that Momentum SGD is the most suitable optimal method for the model, so this method is used in subsequent experiments. Momentum is a critical parameter in Momentum SGD. Therefore, a suitable momentum set could avoid a local minimum of loss (MAE/MAPE), which means we have more probability of building a model having the best performance. Table \ref{table:compare-momentum} compares the influences of several different momentum parameter settings on loss. The results display that MAE and MAPE have minimum values when the momentum is 0.9.

\begin{table}[tb]
\centering
    \caption{
        Base Model Settings
        \label{table:base-model}
        }
    \begin{tabular}{@{}cc@{}}
    \toprule
                            & \multicolumn{1}{c}{\textbf{Parameters}} \\ \midrule
    Type of model           & Deep Neural Networks model              \\
    Number of hidden layers & 2                                       \\
    Nodes of hidden layers  & 24,24                                   \\
    Optimal method          & SGD with momentum                       \\
    Weight initialization   & Xavier                                  \\
    Learning rate           & 0.01                                    \\
    Evaluation methods      & MAE \& MAPE                             \\ \bottomrule
    \end{tabular}
\end{table}

\begin{table}[tb]
\centering
    \caption{
        Comparison of Optimisers
        \label{table:compare-optimiser}
        }
        \begin{tabular}{@{}ccc@{}}
            \toprule
            \multicolumn{1}{c}{\textbf{Optimizer}} & \multicolumn{1}{c}{\textbf{MAE}} & \multicolumn{1}{c}{\textbf{MAPE}} \\ \midrule
            SGD with momentum                      & 4.9072                           & 0.1828                            \\
            ADAM                                   & 4.9407                           & 0.1921                            \\
            SGD                                    & 5.2105                           & 0.2088                            \\
            Adadelta                               & 5.2521                           & 0.1975                            \\
            RMSprop                                & 8.3171                           & 0.4474                            \\ \bottomrule
        \end{tabular}
\end{table}

\begin{table}[tb]
\centering
    \caption{
        Comparison of Momentum
        \label{table:compare-momentum}
        }
        \begin{tabular}{@{}ccc@{}}
            \toprule
            \multicolumn{1}{c}{\textbf{Momentum (SGD)}} & \multicolumn{1}{c}{\textbf{MAE}} & \multicolumn{1}{c}{\textbf{MAPE}} \\ \midrule
            0.9                                         & 4.9072                           & 0.1828                            \\
            0.5                                         & 5.0804                           & 0.189                             \\
            0.99                                        & 5.7209                           & 0.2218 \\ \bottomrule
        \end{tabular}
\end{table}

\begin{table}[tb]
\centering
    \caption{
        Comparison of the Number of Hidden Layers
        \label{table:compare-hidden-layers}
        }
        \begin{tabular}{@{}ccc@{}}
            \toprule
            \multicolumn{1}{c}{\textbf{No. of hidden layers}} & \multicolumn{1}{c}{\textbf{MAE}} & \multicolumn{1}{c}{\textbf{MAPE}} \\ \midrule
            2                                                    & 4.9072                           & 0.1828                            \\
            1                                                    & 5.4508                           & 0.199                             \\
            3                                                    & 4.9347                           & 0.1933                            \\
            4                                                    & 5.7123                           & 0.2663  \\\bottomrule
        \end{tabular}
\end{table}

\begin{table}[tb]
\centering
    \caption{
        Comparison of Hidden Size
        \label{table:compare-hidden-size}
        }
        \begin{tabular}{@{}ccc@{}}
            \toprule
            \textbf{Hidden layers' size} & \textbf{MAE} & \textbf{MAPE} \\ \midrule
            24,24                               & 4.9072       & 0.1828        \\
            24,12                               & 5.4508       & 0.199         \\
            24,48                               & 4.6795       & 0.1805        \\
            24,72                               & 5.033        & 0.1833        \\
            12,24                               & 5.2832       & 0.1942        \\
            48,24                               & 4.9842       & 0.1991        \\ \bottomrule
        \end{tabular}
\end{table}

\begin{table}[tb]
\centering
    \caption{
        Comparison of Weight Initialisation Method
        \label{table:compare-weight-intiialisation}
        }
        \begin{tabular}{@{}ccc@{}}
            \toprule
            \textbf{Weight initialization} & \textbf{MAE} & \textbf{MAPE} \\ \midrule
            Xavier                         & 4.9072       & 0.1828        \\
            He                             & 4.9622       & 0.1917        \\
            Random                         & 5.1104       & 0.202   \\ \bottomrule
        \end{tabular}
\end{table}

\begin{table}[tb]
\centering
    \caption{
        Comparison of Learning Rate
        \label{table:compare-learning-rate}
        }
        \begin{tabular}{@{}ccc@{}}
            \toprule
            \textbf{Learning rate} & \textbf{MAE} & \textbf{MAPE} \\ \midrule
            0.01                   & 4.9072       & 0.1828        \\
            0.05                   & 4.9693       & 0.1989        \\
            0.005                  & 5.0926       & 0.1966        \\ \bottomrule
        \end{tabular}
\end{table}

\begin{table}[tb]
\centering
    \caption{
        Overall Comparision of Model Results
        \label{table:compare-model-results}
        }
        \begin{tabular}{@{}ccc@{}}
            \toprule
            \textbf{Model}    & \textbf{MAE} & \textbf{MAPE} \\ \midrule
            Best model (DNN)  & 4.6795       & 0.1805        \\
            SVM               & 8.1791       & 0.5626        \\
            Linear Regression & 7.8101       & 0.4954        \\
            Random Forest     & 7.7893       & 0.6108        \\
            XGBoost           & 15.8476      & 0.8998        \\ \bottomrule
        \end{tabular}
\end{table}

\subsubsection{Optimizer and Parameters}
A hidden layer of DNN is proposed to solve nonlinear problems. We usually add a small number of hidden layers for simple nonlinear problems (data structure is simple) and use more hidden layers for computer vision and natural language processing tasks. Theoretically, the fitting ability of the model would be improved as the number of hidden layers increases. However, models with more hidden layers would lead to overfitting, increased training time, and difficulty in convergence for simple tasks. Therefore, we only set up a comparative experiment of one, two, three and four hidden layers (the number of nodes in each hidden layer is 24) to find the best-hidden layer number setting. Table \ref{table:compare-hidden-layers} presents that the model with two hidden layers has the smallest MAE and MAPE, and loss would be raised after adding more hidden layers.

The number of hidden nodes is a significant parameter because too few and too many nodes could lead to the under-fitting and overfitting of the model. Too few nodes make the model unable to fully learn the information in the data, leading to underfitting. On the other hand, when a neural network has too many nodes, the limited amount of information in the training set is insufficient to train all the neurons in the hidden layer, which would cause overfitting. Table \ref{table:compare-hidden-layers} has proved that the model with two hidden layers has the best performance, so we conducted a comparison test of Table \ref{table:compare-hidden-size} based on this conclusion. 6 different pairs of nodes in two hidden layers and their related evaluation result have been shown in Table \ref{table:compare-hidden-size}. It is obvious that under a fixed node (24) in the first hidden layer, the model with 48 nodes in the second hidden layer has achieved the minimum MAE 4.6795 and MAPE 0.1805, compared to other candidate nodes (12, 24, 72) of the second hidden layer; in addition, the model with 24 nodes in the first hidden layer has achieved the minimum MAE 4.9072 and MAPE 0.1828, compared to other candidate nodes (12, 48) of the first hidden layer when nodes of the second hidden layer fixed 24. Overall, the model performs best when the number of nodes in the two hidden layers is 24 and 48, respectively.

\subsubsection{Weight Initialization}
Random initialisation is the commonly used method of weight initialisation. However, this method could cause the initial weight to be too large or too small, leading to weight vanishing or exploding in the gradient descent process. Xavier and He weight initialisation can be used to tackle this problem. Table \ref{table:compare-weight-intiialisation} compares models using three different weight initialisation methods. The results display that model with Xavier initialisation has the smallest MAE of 4.9072 and MAPE of 0.1828.

\subsubsection{Learning Rate}
The learning rate is a parameter that determines the convergence speed of the model. A larger learning rate causes the model difficult to converge, while a lower learning rate would cause a local minimum and slow convergence speed. Therefore, setting an appropriate learning rate is crucial in the model training process. Table \ref{table:compare-learning-rate} presents the model's performance with three different learning rate settings. The results illustrate that the model with a 0.01 learning rate has achieved a minimum MAE of 4.9072 and MAPE of 0.1828.

\subsubsection{Model comparison}
According to results highlighted in Table \ref{table:compare-model-results}, the best model is achieved by the DNN model with momentum SGD (momentum = 0.9, learning rate = 0.1), two hidden layers (24 nodes and 48 nodes in the first and second hidden layer respectively) and Xavier initialisation.
In order to prove the importance and superior performance of our model, we compared it with some other common and useful machine learning methods, including SVM, Linear Regression, Random Forest and XGBoost. Table 8 presents that the DNN model, as a pre-trained model built, performs much better in the source domain than other machine learning algorithms.

\begin{figure}[tb]
    \centering
    \includegraphics[width=0.65\linewidth]{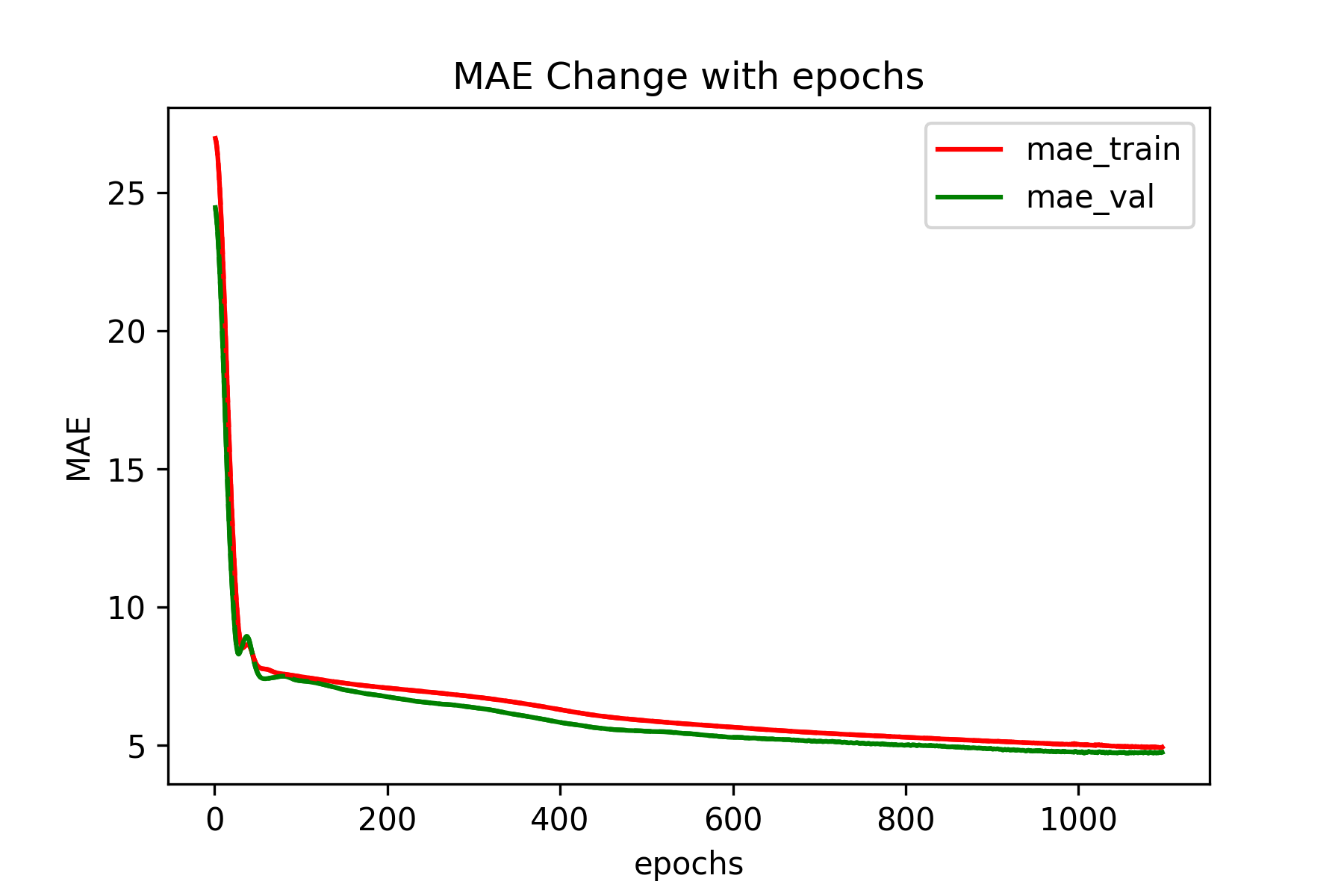}
    \caption{The convergence process of MAE on train and test data
    \label{fig:mae-convergence}}
\end{figure}

\begin{figure}[tb]
    \centering
    \includegraphics[width=0.65\linewidth]{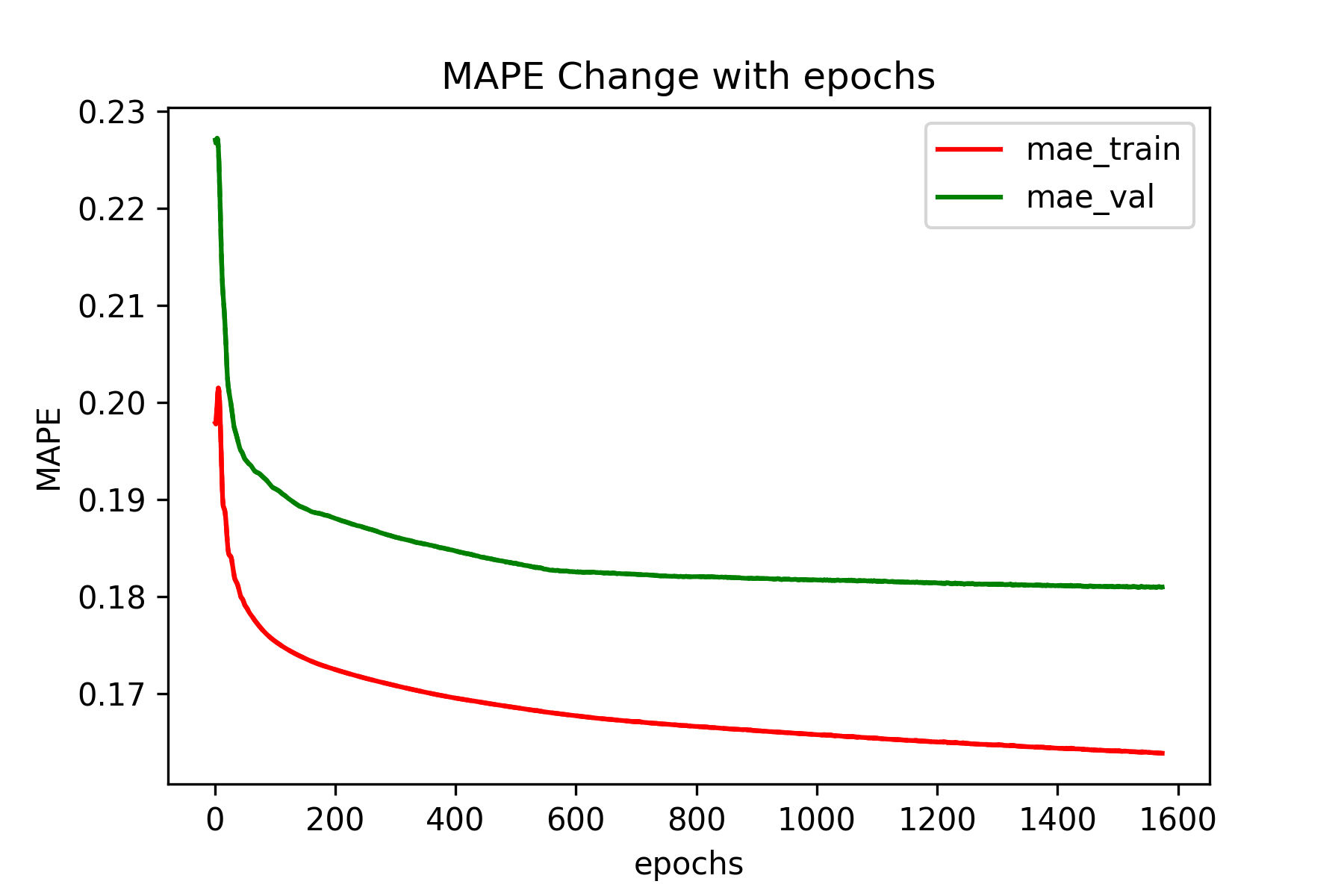}
    \caption{The convergence process of MAPE on train and test data
    \label{fig:mape-convergence}}
\end{figure}

In addition, the number of epochs represents the training schedule, which is the number of times the data has been trained. Figure \ref{fig:mae-convergence} and Figure \ref{fig:mape-convergence} compares the performance of different training schedules; the number of epochs ranges from 0-1000. The MAE and MAPE converge quickly when the epoch is less than 150 and converge slower after 150, finally converging completely. Figure 10 shows the trend of MAE and MAPE on the training and testing sets as the number of epochs increases. The comparison shows that MAE performs similarly on the training and testing sets (the two lines almost overlap), so it is better to choose MAE as the evaluation metric.

\subsection{In Source Domain}
Transfer learning optimises the model by sharing the parameters of the pre-trained model with the target model by mapping the source and target domains to the same space and minimising the distance between them
\begin{itemize}
\item The output layer of the pre-trained model is removed to obtain the weights of the source domain model, and the input and hidden layers are retained.
\item A linear layer is added, and the dataset from the target domain is put in the model. Since the target domain is also regression, the linear layer has one node.
\item The model has been trained with the target dataset in the target domain.
\end{itemize}
The Target dataset has the same feature space as the source domain dataset. The transfer learning model uses these features to predict the PLI of the ports. To highlight the performance of the transfer learning model, SVM, Linear Regression, Random Forest and XGBoost model were compared.
As shown in Table \ref{table:compare-dnn-and-traditional-model-results}, the transfer learning model based on DNN has the smallest MAE and MAPE, which are less than 0.5 and less than 0.03, respectively, while all the other machine learning models have similar results, with MAE between 4-5 and MAPE between 1-3. Since the results are different for each training process, the best model (DNN) evaluation metric is a range of values. After more than 50 replicate experiments, it is found that the MAE of the best model is always less than 0.5, and the MAPE is always less than 0.03, so we use a range of values to represent it. Therefore, the prediction error of the other models is much larger than the transfer learning model based on DNN. This result shows that transfer learning is effective for predicting PLI and can be applied to small datasets.

\begin{table}[tb]
\centering
    \caption{
        Overall Comparison of Transfer Learning-based DNN Against Traditional Models
        \label{table:compare-dnn-and-traditional-model-results}
        }
        \begin{tabular}{@{}ccc@{}}
            \toprule
            \textbf{Model}    & \textbf{MAE}  & \textbf{MAPE} \\ \midrule
            Best model (DNN)  & $<$ 0.5       & $<$ 0.03      \\
            SVM               & 4.8686 ± 2.38 & 2.1954        \\
            Linear Regression & 4.2845 ± 3.37 & 3.0921        \\
            Random Forest     & 4.6717 ± 0.96 & 2.2294        \\
            XGBoost           & 4.2350 ± 4.03 & 1.5657    \\ \bottomrule
        \end{tabular}
\end{table}

\section{Conclusion}
The common approach for measuring marine pollution is the conventional method of collecting water samples or sediment and conducting analysis in the laboratory. However, these methods are accurate but time-consuming, geographically constrained and require professional resources. In order to solve the drawbacks of this common approach, this research aims to utilise the economic activities and ecological factors data of a seaport area as a foundation and then forecast the level of marine pollution the area contains. The prominent advantage of this approach is cost-effective, eliminating the cost of water sampling, sediment sampling and professional analysis. Meanwhile, the data on economic activities and ecological factors are easily accessible from the statistical bureau and weather websites.
In this research paper, we have successfully built a DNN-based transfer learning model to predict the PLI of water quality in 6 ports in NSW, Australia. Experiment results show that the DNN-based transfer learning model performs better than traditional machine learning models. Furthermore, the error MAE is less than 0.5, and the MAPE is less than 3\%, which means that the model we constructed could be well applied to predicting PLI in the ports of NSW, Australia.
In addition, industries such as tourism, aquaculture and marine life protection could be benefited from this approach. As if tourists, divers especially, have the intention of exploring coastal waters around a specific NSW area, and want to know the water quality in advance, what tourists could do is collect relevant data such as population density, temperature, wind speed, etc. input these data into the model from this project to result in a range of PLI, so that makes it convenient for decision making. This research has outstanding economic and time advantages compared to traditional water quality analysis methods. Compared with the standard machine learning model to predict water quality, the transfer learning method proposed in this study has higher prediction accuracy and only requires less data.

\vspace{6pt}

\printbibliography

\end{document}